\documentclass[10pt]{article} % For LaTeX2e
\usepackage[preprint]{rlc}

\usepackage{amssymb}      % Defines common symbols like \mathbb R
\usepackage{mathtools}     % Extends amsmath, providing common math tools
\usepackage{mathrsfs}      % Enables \mathscr, which can work in cases that \mathcal does not
\usepackage{graphicx}      % For including images
\usepackage{subcaption}     % Allows for the use of subfigures and subcaptions
\usepackage[space]{grffile}   % For spaces in image names
\usepackage{url} 
\usepackage{times}
\begin{document}

% paper title
\title{Towards Interpretable Foundation Models of Robot \\ Behavior: A Task Specific Policy Generation Approach
\vspace{-5.5mm}}

% You will get a Paper-ID when submitting a pdf file to the conference system
% \author{Author Names Omitted for Anonymous Review. Paper-ID [add your ID here]}

\author{Isaac Sheidlower  \\
    Isaac.Sheidlower@tufts.edu \\
    School of Engineering \\
    Tufts University
    \And
    Reuben Aronson \\
    Reuben.Aronson@tufts.edu \\
    School of Engineering \\
    Tufts University   
    \And
    Elaine Schaertl Short \\
    Elaine.Short@tufts.edu \\
    School of Engineering \\
    Tufts University}

\maketitle
\vspace{-9mm}
\begin{abstract}
\vspace{-5mm}
Foundation models are a promising path toward general-purpose and user-friendly robots. The prevalent approach involves training a ``generalist policy'' that, like a reinforcement learning policy, uses observations to output actions. Although this approach has seen much success, several concerns arise when considering deployment and end-user interaction with these systems. In particular, the lack of modularity between tasks means that when model weights are updated (e.g., when a user provides feedback), the behavior in other, unrelated tasks may be affected. This can negatively impact the system's interpretability and usability. We present an alternative approach to the design of robot foundation models, Diffusion for Policy Parameters (DPP), which generates stand-alone, task-specific policies. Since these policies are detached from the foundation model, they are updated only when a user wants, either through feedback or personalization, allowing them to gain a high degree of familiarity with that policy. We demonstrate a proof-of-concept of DPP in simulation then discuss its limitations and the future of interpretable foundation models.
\end{abstract}

\vspace{-6mm}
\section{Introduction}
\vspace{-4mm}
 Current efforts in creating task-generalizable, novice-friendly robots are largely focused on foundational models of robot behavior. The goal of such a model is to have a user ask the robot to do an arbitrary task via verbal or non-verbal communication, then have the robot perform the task with little to no further human supervision or intervention. The relatively few robot foundation models that exist can all be categorized as ``generalist robot policies.'' In particular, the input is an observation of the robot's state combined with a language or goal embedding that specifies the task, and the output is a robot action, such as end-effector displacement. This is the same input-output relationship as a typical task-specific Reinforcement Learning (RL) policy. 
 
 However, these generalist policies have limitations that may pose serious problems when actually deployed for users. In deployment, a foundation model should be able to learn from data in many different environments and tasks and personalize to individual users in response to training. However, generalist policies are not localized with respect to task: new feedback for one task could change model behavior in a completely unrelated task. This property limits the ability of users to personalize the system or learn what to expect of its behavior for a given task. In this work, we propose an alternative approach where the foundation model is a policy generator, which outputs standalone, task-specific policies. We discuss potential benefits of this approach to robot foundation models as well as potential challenges with generalist policies as a sole solution.

We present Diffusion for Policy Parameters (DPP), a method for generating standalone task policies conditioned on a task specification. We present a proof-of-concept implementation for smaller grid-world tasks. We show, to the best of our knowledge, for the first time that one can learn representations to generate policies directly in parameter space, without the need for policy search. Finally, we discuss limitations of the DPP approach and how to both potentially resolve them and other alternative methods towards generating task-specific policies. Our results show that DPP is a promising approach to robot behavior foundation models and warrants further investigation. 

% These best performing of these models typically make use of a language-perception model, such as CLIP, to combine in an embedding language instructions for the task in an image or video of the current robot. They also consist of a large behavioral model, typically a transformer, which takes in such an embedding as input and outputs a sequence of actions for the robot to take. In addition to the large amounts of data the vision-language model was trained on, there is a continual effort to make multi-robot and multi-scene data sets from across academia and industry alike. Like the large amounts of data used to train these models, the models themselves are both large in size, making them costly to deploy, and potentially, more difficult to interpret/explain and adjust in real-time. For example, if one wanted to use such a large behavior policy in Shared Autonomy for example, one may need to query the policy for prompts simultaneously and at a high frequency. In other words, it is important to consider how foundational behavior models are or are not compatible with these various HRI applications, as well as, how we can approach designing and building foundation models which are both compatible with a wide range of HRI applications, while still being highly performant. 

% Instead of trying to make a large generalist policy, we should shift our focus to a large generalist small policy generator. 
% In other words, we want a foundational model for generating small policies that are faster and cheaper to run inference, and potentially more interpertable.

\vspace{-5mm}
\section{Related Work}
\vspace{-4mm}
Recent technological advancements in Generative AI (Gen-AI) including the transformer architecture \citep{vaswani_attention_2023}, and diffusion models \citep{ho_denoising_2020}, that are both high performing and scale to large amounts of data, have allowed for the development of larger and more general purpose task models. In the case, of language prediction, Large Language Models (LLMs) such as GPT-4 \citep{openai_gpt-4_2024} and LLaMa \citep{touvron_llama_2023}, have enabled a single large-scale model to perform a variety of linguistic tasks \citep{zhao_survey_2023}. Similarly, multi-model models such as CLIP \citep{radford_learning_2021} and Latent Diffusion Models (LDMs) \citep{rombach_high-resolution_2022}, have allowed for similar generalizability in image-language embedding and language-conditioned image generation respectively. These technologies are starting to be applied for general robotic manipulation. Most notable are the large robot behavior policies RT-1/2-X \citep{collaboration_open_2024,brohan_rt-2_2023} and Octo \citep{ghosh_octo_nodate}, and diffusion architectures for fast imitation learning across a variety of tasks \citep{chi_diffusion_2024, ze_3d_2024}. These models are policies which transform observations to actions, and, while these models are promising and are beginning to see impressive success across many tasks, there is little modularity between learning and executing across different tasks. 

% In particular, a user teaching or personalizing a generalist policy for one task, may have unwanted down stream consequences on other tasks. Due to this concern, we propose other methods and evaluations for robot foundation models be explored with user-interaction in mind. 

RL has previously had success for both robot manipulation \citep{ibarz_how_2021, nguyen_review_2019} and being applied to human-robot-interaction (HRI) scenarios \citep{reddy_shared_2018, akalin_reinforcement_2021, park_model-free_2019}. When a user interacts with a consistent and self-contained policy, they can better predict its behavior \citep{cruz_explainable_2023, horter_varying_2023}, customize and personalize its behavior through feedback \citep{bobu_feature_2021, arzate_cruz_survey_2020, brawer_interactive_2023, sheidlower_online_2024} and learn to leverage its dynamics to accomplish novel tasks \citep{aronson_intentional_2024, gopinath_human---loop_2017}. This work on how users interact with RL systems highlight the value of explainable and modular policies for positive outcomes.

\begin{figure*}[t]
\centering
  \vspace{-8mm}
 \includegraphics[scale=.1, width=.85\textwidth]{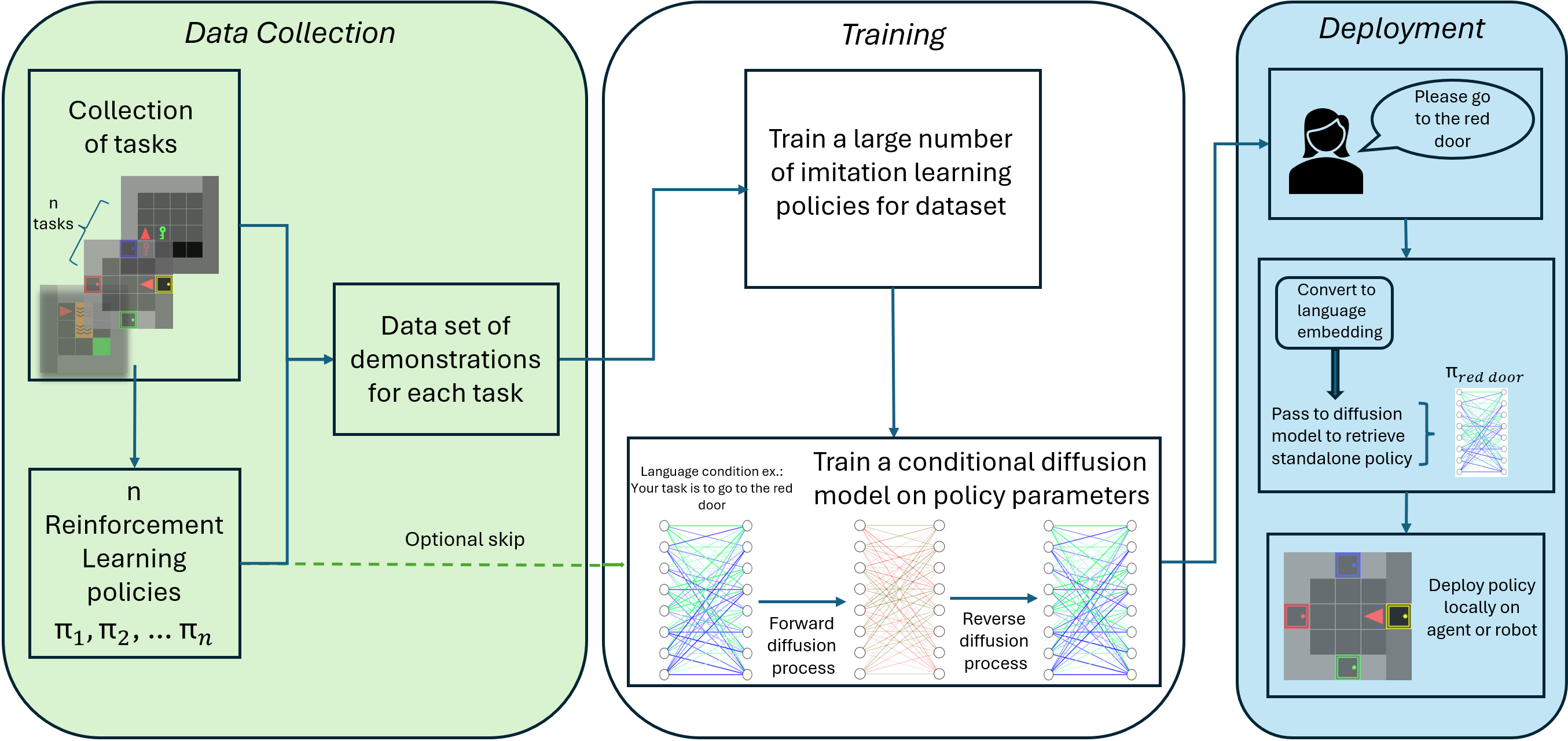}
 \vspace{-3mm}
 \caption{The DPP foundation model of robot behavior design approach}\label{fig:study}
 \vspace{-6mm}
\end{figure*}

\vspace{-5mm}
\section{Potential Challenges with Robot Foundation Models as Generalist Policies}
\vspace{-4mm}
% Creating a generalist robot policy that can function across different tasks and robots is a common and intuitive approach to making general purpose robots. Furthermore, this approach has shown impressive task success and generalizability. However, when we consider end-users interacting with robots that use this approach, concerns around the systems' explainability and consistency arise. 

Robots should be able to learn from feedback and have real-time behavior personalization for any given task. If the policy the user is interacting with is a generalist robot policy, two problems may limit a user's ability to do this. The first is that when a user teaches the robot a new task or personalizes the behavior for a certain task, the behavior in separate and unrelated tasks may be affected. This may jeopardize the interpretability and legibility of the system \citep{bobu_aligning_2024}. Another is that updates to the base of the model from the organization which developed the model may have downstream affects on specific tasks/robot behavior that may be unnexpected or undesired by a user. This is already the case with consumer-available LLMs such as ChatGPT, however, in the case of robotics, the consistency of the robot's behavior is a crucial component to the user's ability to teach and interact with the robot. In fact, robots spontaneously acting in unexpected ways around users may cause physical safety concerns beyond those posed by systems operating solely on language. Thus, making sure that a robot's task behavior is changed when and how a user wants is crucial.

\vspace{-5mm}
\section{Diffusion for Policy Parameters (DPP)}
\vspace{-4mm}
We present Diffusion for Policy Parameters (DPP), a novel approach for learning how to generate standalone policies for individual tasks. DPP alleviates some of the concerns mentioned in the prior section. We then present a proof-of-concept evaluation in a grid-world simulation. This is, to the best of our knowledge, the first generative approach for creating policies in parameter space. While policy search \citep{plappert_parameter_2018, taylor_transfer_2007, levine_guided_2013, kalyanakrishnan_empirical_2009} and exploration over policy parameters \citep{fontaine_quality_2021, mouret_illuminating_2015, tjanaka_pyribs_2021} have been explored, generative AI techniques have not been used directly in parameter space.

The DPP method (Figure 1.) learns a conditional diffusion model for generating policies in in policy parameter space. The steps for DPP are: collect a dataset of language/goal conditioned tasks and a dataset of demonstrations over those task; train a large set of policies on either the demonstrations or tasks themselves; then train a diffusion model conditioned on the task description and takes the parameters of the policies as input. The result is a model that leads to an interaction similar to a generalist robot policy: a user asks for or demonstrates a task, and then the robot autonomously executes that task, with the option of further human-in-the-loop fine-tuning if necessary. The key difference being in DPP, a policy independent of the foundation model is generated to execute the task. To study whether DPP is a viable approach for learning to generate policies, we must show it can lead to a model which conditionally generates ``good'' policies.

\textbf{4.1, Environment and Data Collection} We ran experiments in the Minigrid environment \citep{chevalier-boisvert_minigrid_2023} for its suite of language-conditioned tasks on which we can train many agents on in a relatively small amount of time and with limited hardware. All tasks have a similar reward structure: sparse reward with a time-step penalty, resulting in a cumulative reward between 0 and 1. To generate a large number of tasks, we took three language-conditioned tasks and made each goal specification within that task its own task. In particular, we took the environments Fetch, Go to Door, and Go to Object, and for each possible object configuration, made that a task (e.g. Go to Door contains both the ``go to red key'' and ``go to blue box'' task specification, and we treat each as a task to train an agent on). We chose the 5x5 versions of each task for computational efficiency and quicker training. We then collected many seeds for each task to ensure random goal positions and obstacles, resulting in 84 unique tasks. While these tasks are significantly simpler than in-the-wild robot tasks, they provide a wide range of separate policies to train on. %In the real-robotics case, tasks would range from ''make coffee" to ''clean the windows" or ''water the flowers."

To collect policy data on these tasks, we trained 84 RL agents using PPO \citep{schulman_proximal_2017} to optimality (achieving a mean reward $>.98$). We then trained behavior cloning (BC) agents on trajectories collected from the RL agents until they received a near-optimal average reward of $>.85$. We chose BC as opposed to RL for every agent because it was more timely to train and collect the policies. We trained approximately 1000 agents for each of these tasks, discarding tasks where BC did not achieve high reward given the allotted trajectories. This resulted in BC agents for 64 of the 84 tasks and resulted in 74,000 trained policies. 
% This process took about 72 hours in total on a consumer-grade computer with a single 4070 TI Super. 

\begin{table}[t]
\vspace{-8mm}
\scalebox{.8}{
\begin{tabular}{|l|l}
\cline{1-1}
\small
\textbf{DPP Model Architecture} &                                                                            \\ \hline
Language Embedding       & \multicolumn{1}{l|}{bge-small-en-v1.5 (size: 384) \citep{xiao_c-pack_2023}}                                                  \\ \hline
Noise Schedule         & \multicolumn{1}{l|}{Cosine, 1000 steps \citep{nichol_improved_2021}}                                                        \\ \hline
Noise Type           & \multicolumn{1}{l|}{Gaussian \citep{ho_denoising_2020}}                                                             \\ \hline
Model Architecture       & \multicolumn{1}{l|}{\begin{tabular}[c]{@{}l@{}}Transformer \citep{vaswani_attention_2017}, 48 heads, 12 depth, 768 width\end{tabular}}                      \\ \hline
Batch Size           & \multicolumn{1}{l|}{128}                                                               \\ \hline
Input/Output Dimensions     & \multicolumn{1}{l|}{\begin{tabular}[c]{@{}l@{}}32x82; 32 for MLP policy hidden layer, 82 = 75 (observation size) + 7 (action size)\end{tabular}} \\ \hline
\end{tabular}}
\vspace{-3mm}
\caption{DPP model architecture used in experiments}
\vspace{-4mm}
\end{table}

\textbf{4.2, Model Design} Given the dataset of policies, we trained a conditional diffusion model which takes as input a language description of the task and outputs an end-to-end policy network for that task. The model architecture and description can be found in Table 1. The architecture was largely decided on based on trial and error. However, two key decisions were necessary to effectively learn in parameter space. The first was to use an entirely transformer-based architecture, as opposed to, e.g., a U-Net architecture \citep{ronneberger_u-net_2015, ho_denoising_2020}. The other was to use the hybrid loss as proposed in \citep{nichol_improved_2021}. We also experimented with various loss functions based on evaluation of the generated policies, but they did not lead to high performance.

\begin{table*}[t]
\resizebox{\textwidth}{!}{%
\begin{tabular}{l|l|l|l|l|l|l|l|l|}
\cline{2-9}
 &
 \begin{tabular}[c]{@{}l@{}} Diffusion \\ Sample \\ Policy\end{tabular} &
 Random Policy &
 \begin{tabular}[c]{@{}l@{}}Training \\ Parameters \\ (TP) Mean\end{tabular} &
 TP Median &
 TP Mode &
 \begin{tabular}[c]{@{}l@{}}Mixture of \\ Samples \\ (MoS), m=4\end{tabular} &
 \begin{tabular}[c]{@{}l@{}}(MoS), \\ m=8\end{tabular} &
 \begin{tabular}[c]{@{}l@{}}(MoS), \\ m=16\end{tabular} \\ \hline
\multicolumn{1}{|l|}{Avg. Return} &
 \textbf{0.766}±0.16 &
 \textbf{0.198}±0.14 &
 \textbf{0.189}±0.14 &
 \textbf{0.205}±0.12 &
 \textbf{0.125}±0.16 &
 \textbf{0.816}±0.19 &
 \textbf{0.878}±0.15 &
 \textbf{0.886}±0.16 \\ \hline
\end{tabular}%
}
\vspace{-3mm}
\caption{Results from experiments}
\vspace{-8mm}
\end{table*}

\textbf{4.3, Evaluation and Results} The evaluation results of the final trained model can be found in Table 2. The evaluation aims to show the model generates meaningful policies in parameter space. For each baseline, we took average performance across all 64 tasks, with 10 runs each on random seeds. ``Diffusion Sample Policy'' refers to a single sample from the diffusion model conditioned on the task description. We primarily compare to baselines as a means to ensure the model is not learning trivial local minima. If it is not, then we expect a single sampled policy to significantly outperform the baselines. "Random Policy" refers to an agent that takes a random action in each state. The "Mean," "Median," and "Mode" baselines refer to taking those operations on all of the parameters in the dataset for the specified task. The sample policy significantly outperforms all baselines indicating that the model is learning to generate meaningful and performant polices. A single sample, however, achieves slightly lower returns than the agents in the training set. To achieve a similar performance, we take a simple mixture approach where we sample $n$ policies, and for each observation, take the most common output action. This is referred to as Mixture of Samples (MoS) in Table 2. 

\textbf{4.4, Limitations}
Despite promising early results, there are key limitations with the evaluation regarding extrapolating the results to real-world robots. Though diffusion models and transformers have been shown to scale well with large amounts of robot data, we have not shown this scalability with policy parameter space learning. Similarly, we emphasize that the training data needed for DPP is different than for a generalist policies: DPP requires a dataset of trained policies (which could be gathered through simulation or a cross organization effort similar to the Droid dataset \citep{khazatsky_droid_2024}), rather than a demonstration dataset. However, for DPP to scale and generalize across tasks, it will likely need both a policy and a demonstration dataset. We have also only demonstrated results in an environment with a discrete action space. Although some generalist policies, such as \citep{ghosh_octo_nodate}, have had issues with discrete action spaces, we believe a robot foundation model should be able to handle both discrete and continuous actions. These limitations warrant further investigation and to be addressed in future work.

\vspace{-5mm}
\section{Discussion}
\vspace{-4mm}
While generalist robot policies as robot behavior foundation models show clear successes, they do not maintain properties of locality and explainability that would be desired for a deployed system. To limit these concerns, we presented DPP, an alternative which may alleviate some of the outlined concerns. DPP generates smaller, standalone policies for each task; this approach means that those policies are not affected by a user teaching the robot other tasks or by unwanted updates to the general foundation model. 

 % allowing a user to gain a high degree of familiarity with those policies, 

% rma added -- put all the HRI perspective stuff you want here
Enabling policies to be stable and therefore more predictable is a key feature for human-usable robots. Human-robot interaction research has consistently shown that robot models need to be not just performant, but also predictable \cite{lichtenthaler_legibility_2016}. A predictable robot system not only improves its interpretability, but also allows a user to gain a high degree of familiarity with those policies, and in turn use them to accomplish novel tasks. In this work, we embed this predictability and usability directly into the structure of the model without compromising its flexibility to learn from new data. With foundation models in their infancy, it is an ideal time to explore how these powerful generalized models can be made more usable.

Future work could explore other methods to make foundation models more stable and usable, especially by allowing the user to choose when and how a task policy should be updated. For example, a robot might come deployed with a suite of policies for very common tasks, with the capacity to learn new tasks from the user through human-in-the-loop learning \citep{ravichandar_recent_2020, liu_robot_2023}. Another approach is to use sim-to-real RL to train new policies when needed by the user. For example, Eureka \citep{ma_eureka_2024} uses LLMs and an iterative training procedure to design reward functions for arbitrary tasks. This approach has similar benefits as DPP, but depends on having an accurate simulator and may not be responsive enough for users, as the robot needs to learn tasks from scratch when the user requests it. An advantage, however, is that since it uses a task-specific reward function, it may be more explainable. 

 % allowing a user to gain a high degree of familiarity with those policies, 

We primarily focused on improving interpretability and modularity relative to generalist policies, but there are also other exciting directions for future research towards usable generalist policies. For example, work is needed on how to explain the behavior of a generalist robot policy. Explainable AI techniques for large models are constantly improving, but more work is needed to understand how techniques for explaining robot behavior apply to generalist robot policies: much prior work in explainable robotics and AI either assumes the robot was trained with RL (see \citep{milani_explainable_2024} for a taxonomy of explainable RL) or requires semantic knowledge of its past interactions \citep{setchi_explainable_2020}. Some approaches, such as directly generating explanations for non-interpretable policies are easily applicable to generalist robot policies, while others, such as generating intrinsically interpretable policies, may not be. 

% Another concern is with the hardware-scalability of these generalist policies. In robotics it is often necessary for weights of policies to be stored locally to run inference on and for the user to interact with. If a large generalist policy is stored on a remote server, because the input to the policy is observations, each robot will query that policy at a relatively high frequency, especially in human-robot collaboration scenarios or safety critical (saftey-critical robots in dynamic settings often have controllers that run at $\geq$60hz for example \citep{nguyen_robust_2022, molnar_model-free_2022}). Inference frequency requirements may become unreasonably high as robots become increasingly prevalent. 

\vspace{-5mm}
\section{Conclusion}
\vspace{-4mm}
In this work, we discussed key areas for improving the paradigm of generalist robot policies as robot behavior foundation models. To better empower end-users, these models need better modularity and independence between tasks, to support a user's ability to understand and leverage the system's dynamics, independent of weight changes in the foundation model. Towards this goal, we presented the Diffusion for Policy Parameters (DPP) approach to a robot behavior foundation. Our preliminary results show that this is a promising method for generating task-specific, standalone policies conditioned on a task specification. As the research community explores methods such as DPP, we can ensure the future of foundational robot-behavior models empower end-users with a high degree of understanding and control over the robot.

\section*{Acknowledgments}
The work described here was supported in part by the US
National Science Foundation (IIS-2132887)
%% Use plainnat to work nicely with natbib. 

% \bibliographystyle{plainnat}
% \bibliography{references}
\bibliography{references}
\bibliographystyle{rlc}

\end{document}